\documentclass[preprint,12pt]{elsarticle}

\usepackage{setspace}
\usepackage{multirow}
\let\today\relax
\makeatletter
\def\ps@pprintTitle{%
    \let\@oddhead\@empty
    \let\@evenhead\@empty
    \def\@oddfoot{\footnotesize\itshape
         {Preprint} \hfill\today}%
    \let\@evenfoot\@oddfoot
    }
\makeatother

\usepackage{amssymb}

\usepackage{lineno}
\usepackage{hyperref}

\usepackage{pgfplots}
\usepackage{pgf, tikz}
 
\usetikzlibrary{arrows, automata}
\usetikzlibrary{shapes.multipart}
\usetikzlibrary{quotes,arrows.meta}%
\usetikzlibrary{decorations.pathreplacing}
\usetikzlibrary{positioning}
\usetikzlibrary{matrix}
\usetikzlibrary{backgrounds}
\usetikzlibrary{fit, patterns}
\usepackage[utf8]{inputenc}
\usepgfplotslibrary{groupplots,dateplot}
\usetikzlibrary{shapes.arrows}
\pgfplotsset{compat=newest}
\usepackage{graphicx}
\usepackage{graphbox}
\usepackage{color}
\usepackage{ifthen}
\usepackage{bm}
\usepackage{xurl}
\usepackage{adjustbox}
\usepackage{subcaption}
\usepackage{comment}
\usepackage{multirow}
\usepackage{xspace}
\usepackage{pgfgantt}

\usepackage{amssymb}
\usepackage{pdfpages}
\makeatletter
\newcommand\saveequation[2]{%
  \@namedef{equation@#1}{#2}%
}
\newcommand\useequation[1]{%
  \@nameuse{equation@#1}%
}
\makeatother
\RequirePackage{amsmath,bm}

\usepackage[figuresright]{rotating}

\usepackage{booktabs}

\usepackage[disable, colorinlistoftodos,textsize=small,textwidth=2.5cm]{todonotes}
\setuptodonotes{backgroundcolor=yellow!10!white}
\newcounter{todoListItems}

\usepackage[normalem]{ulem}

\saveequation{massenergy}{E = mc^{2}}
\saveequation{curlE}{\nabla \times \bm{E} = 0}
\newcommand{\activationFunction}[1]{\ensuremath{g^{[#1]}}}
\newcommand{\featureMap}[1]{\ensuremath{\bm{A}^{[#1]}}}
\newcommand{\filter}[2]{\ensuremath{F^{[#1]}_{#2}}}
\saveequation{layerActivation}{\featureMap{\ell} = \activationFunction{\ell}(\bm{Z}^{[\ell]}) = \activationFunction{\ell}(\featureMap{\ell-1} * \filter{\ell}{})}

\includecomment{optionalQuestions}
\newcommand{\question}[1]{}
\begin{optionalQuestions}
	\renewcommand{\question}[1]{\noindent\textcolor{gray}{\newline  #1 \newline}}
\end{optionalQuestions}

\newcounter{myexample}[section]

\makeatletter
\DeclareRobustCommand\onedot{\futurelet\@let@token\@onedot}
\def\@onedot{\ifx\@let@token.\else.\null\fi\xspace}

\makeatother

\providecommand{\decoRule}{\rule{.8\textwidth}{.4pt}} 

\usepackage{tabularx}
\let\cite\cite

\newcommand{\norm}[1]{\ensuremath{\lVert #1 \rVert}}
\journal{Computer Vision and Machine Learning Journal}
\begin{document}
\newif\ifbone
\bonetrue 

\begin{frontmatter}
%
%
\title{SSTM: Spatiotemporal Recurrent Transformers for Multi-frame Optical Flow Estimation}
%
%
\author{Fisseha Admasu Ferede}
\author{Madhusudhanan Balasubramanian}
\address{Department of Electrical and Computer Engineering, The University of Memphis, Memphis TN 38152\\[-0.2in]}
\begin{abstract}
Inaccurate optical flow estimates in and near occluded regions, and out-of-boundary regions are two of the current significant limitations of optical flow estimation algorithms. Recent state-of-the-art optical flow estimation algorithms are \textit{two-frame} based methods where optical flow is estimated sequentially for each consecutive image pair in a sequence. While this approach gives good flow estimates, it fails to generalize optical flows in occluded regions mainly due to limited local evidence regarding moving elements in a scene. In this work, we propose a learning-based \textit{multi-frame} optical flow estimation method that estimates two or more consecutive optical flows in parallel from multi-frame image sequences. Our underlying hypothesis is that by understanding temporal scene dynamics from longer sequences with more than two frames, we can characterize pixel-wise dependencies in a larger spatiotemporal domain, generalize complex motion patterns and thereby improve the accuracy of optical flow estimates in occluded regions.  We present learning-based spatiotemporal recurrent transformers for multi-frame based optical flow estimation (SSTMs). Our method utilizes \textit{3D Convolutional Gated Recurrent Units} (3D-ConvGRUs) and \textit{spatiotemporal transformers} to learn recurrent space-time motion dynamics and global dependencies in the scene and provide a generalized optical flow estimation. When compared with recent state-of-the-art two-frame and multi-frame methods on real world and synthetic datasets, performance of the SSTMs were significantly higher in occluded and out-of-boundary regions. Among all published state-of-the-art multi-frame methods, SSTM achieved state-of the-art results on the Sintel Final and KITTI2015 benchmark datasets. 
\end{abstract}
\end{frontmatter}


\section{Introduction}
\label{sec:introduction}
In computer vision, optical flow represents a dense flow field comprised of apparent 2D motion of pixel intensities estimated from a sequence of input images. The goal of optical flow estimation algorithms is to estimate the velocity of pixel coordinate transformation invariant to occlusion, motion blur, out-of-boundary regions and flow magnitude at each pixel location using two or more consecutive frames from a scene.  

Horn and Schunck \cite{horn1981determining} was a pioneering algorithm that introduced a variational energy minimization approach to estimate optical flow using a brightness consistency assumption (pixel intensity during pixel coordinate transformation assumed to be constant for a smaller temporal change) and a spatial smoothness constraint for the optical flow field as part of the objective function. Other classical multi-frame methods use a bank of \textit{hand-crafted} motion filters tuned to capture moving patterns and texture characteristics from each image sequence to estimate the direction and magnitude of optical flow fields \cite{heeger1988optical, gautama2002phase}. 

With the recent advancement of deep learning and availability of large datasets with known ground truth, optical flow estimation is reformulated as an end-to-end learning problem without requiring assumptions regarding the characteristics of images and of motion patterns. Motivated by Heeger's \cite{heeger1988optical} approach to use hand crafted spatiotemporal Gabor filters, Teney \textit{et al.} \cite{teney2016learning} developed a learning-based multi-frame optical flow estimation model based on learnable 3D convolutional neural networks (CNN) and signal processing concepts. FlowNet \cite{dosovitskiy2015flownet} introduced a learning based two-frame optical flow estimation method using CNNs wherein a single flow field is estimated from a pair of image frames. Several other learning based two-frame methods improved upon the FlowNet model, namely the coarse-to-fine pyramid network \cite{ranjan2017optical, sun2018pwc}, multiple intermediate flow estimates and warping based brightness error computation \cite{ilg2017flownet}, receptive field guided motion feature extraction networks \cite{sun2018pwc, salehi2023ddcnet} and the 4D all-pairs correlation volume with gated recurrent network \cite{teed2020raft}. 

    \begin{figure}[ht]
        \centering
    \includegraphics[width=1\textwidth]{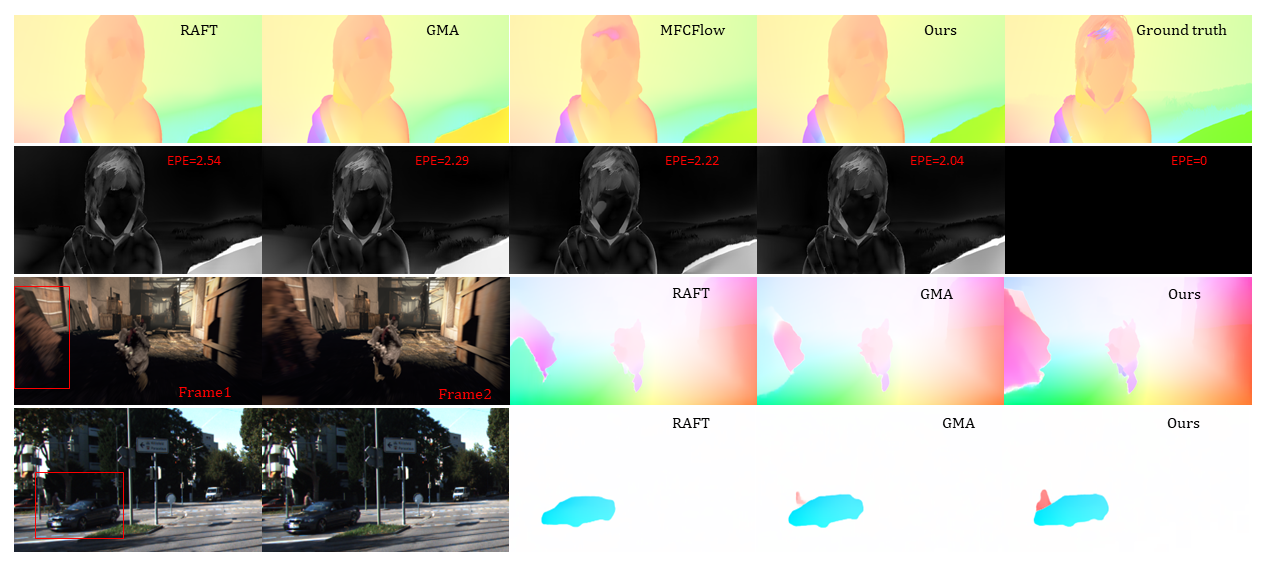}{ 
         \caption{Examples of optical flow estimates (OFEs) using our methods compared with recent state-of-the-art two-frame and multi-frame methods. The regions bounded by red boxes in the input frames represent the regions where our method significantly outperformed the other methods. First and second row show OFEs of sample frames from Sintel Final with \textit{out-of-boundary} scene (bottom right corner) and its corresponding endpoint error map, respectively. The third row shows OFEs of sample frame from Sintel Final pass with large motions (the wing of the dragon captured accurately by our method). The fourth row shows OFEs of sample frames from KITTI2015 dataset where partially \textit{occluded} scene (the person riding a bicycle behind the car) is accurately captured by our method while RAFT \cite{teed2020raft} and GMA \cite{jiang2021learning} failed to do so. }
         \label{fig:comparison-all}}
    \end{figure}

  In dynamic scenes, limited temporal information available from only two frames causes the two-frame methods to have poor generalizations in and near occluded regions as well as in out-of-boundary regions. This is likely because in these regions the flow dynamics are often complex and cannot be fully captured using only two frames. In regions with disappearing scene elements, learning based two-frame methods do not have access to context information as elements appear in only one of the two image frames likely resulting in erroneous flow estimates.

  In this work, we developed a multi-frame optical flow estimation method with the goal of overcoming the limitations of two-frame methods near occluded and out-of-boundary regions. Our methods were designed on the basis of understanding the space-time nature of the input sequence in a wider temporal windows by taking multiple input frames and estimating their flows in parallel to achieve temporal smoothness.  Our model learns any spatiotemporal recurrence structure that can characterize scene dynamics from an image sequence and iteratively estimates optical flows. This gives our method the ability to understand optical flow from a wider non-localized space-time domain when compared to two-frame methods. While we demonstrate our method using three consecutive frames, it can be extended to handle longer sequences with more than three input frames. Figure~\ref{fig:comparison-all} demonstrates optical flow estimates of scenes with occluded and out-of-boundary regions. In these qualitative examples, it can be observed that our methods provides more accurate optical flow estimates than two of the state-of-the-art two-frame methods (RAFT and GMA).

  In general, two-frame methods can be seen as approaches those approximate temporal flow dynamics at each discrete scene location and at each time instant as piece-wise linear geometrical transformation between two consecutive time points or image frames. Our multi-frame optical flow estimation method is aimed at improving the accuracy of these linear transformation estimates between any two consecutive time instants using plausible occlusion information discernible from multiple consecutive frames.  Further, our method provides these estimates at one or more inference time points in parallel thereby forcing the model to learn temporal dependencies among flow estimates.

  We designed our models to estimate optical flow iteratively using 3D convolutional gated recurrent units (3D convGRU) from two or more consecutive images (multi-frame). With the 3D convGRU weights tied among iterations, we facilitated the model to learn any recurrent structure in the image sequence. While we report our training details and experimental observations using three consecutive frames per sequence, our 3D convGRU design allows processing image sequences of arbitrary length. The inputs to the 3D convGRU were aggregated \textit{3D motion features} and spatiotemporal \textit{context features} guided by a space-time \textit{attention mechanism}. The 3D motion features were comprised of i) a 4D correlation volume that exhaustively captured pixelwise similarities among the images (adopted from RAFT \cite{teed2020raft}); ii) intermediate optical flow estimates; and iii) intermediate multi-level flow error estimates obtained by comparing image features of a given image in the sequence with the warped features generated from the next image in the sequence using the intermedial flow estimates. The context features were concurrently extracted from the three consecutive images using a spatiotemporal 3D CNN. Unlike most two-frame methods whose context features are limited to the spatial context of a single image \cite{teed2020raft, jiang2019stm, sui2022craft}, we designed a spatiotemporal context encoder that operates on multiple consecutive images to capture the space-time context from the input sequence. We designed a space-time attention mechanism that captured space-time global dependencies among corresponding locations in the image sequence. A self-attention architecture comprised of two spatial attention networks was used to implement the space-time attention mechanism.


 The structure of this article is organized in the following order. In \textit{Section 2}, we provide a brief review of related work and techniques related to the key features of our methods. In \textit{Section 3}, we discuss the key features as well as the contributions of our proposed multi-frame methods, namely SSTM and SSTM++, and the theoretical basis of our design. In \textit{Section 4}, we present our experiment details, experimental results and a discussion of results based on our observations.  Experimental details presented include the training strategies employed and ablation experiments conducted to elucidate key features of our methods when compared with state-of-the-art methods.  Experimental results presented include both our quantitative and qualitative observations.  In \textit{Section 5}, we conclude this research work with a brief summary of our multi-frame optical flow estimation methods, their performance, and future research directions of this research work.

 The main contributions of this work are as follows:
 \begin{itemize}

    \item We introduce two multi-frame optical flow estimation methods, namely SSTM and SSTM++, those can capture the space-time dependencies among multiple input frames by achieving temporal smoothness to better estimate optical flows in and near occluded regions as well as out-of-boundary regions.

    \item We introduce a new 3D convolutional gated recurrent unit for learning temporal dependency among consecutive flow fields (e.g., to extract missing information in occluded zones at a given time instant from any relevant information from the corresponding zones from nearby time instants) in multi-frame methods; a space-time attention mechanism for extracting attention cues also along the temporal dimension; and a spatiotemporal context feature extraction module.  These three new modules served as backbone of our design.

    \item Our methods outperformed the current state-of-the-art multi-frame method (MFCFlow \cite{chen2023mfcflow}) on both KITTI2015 and Sintel Final benchmark datasets as well as achieved comparable results with current state-of-the-art two frame methods.

\end{itemize}

\section{Related Work}
     \label{sec:related}

     Toshio \textit{et al.} \cite{chin1994probabilistic} developed one of the earliest classical methods for multi-frame optical flow estimation. In this work, the two-frame Horn and Schunck \cite{horn1981determining} optical flow algorithm was extended by introducing temporal smoothness constraint to the objective function using a Kalman filter. The spatial smoothness constraint used in the Horn and Schunck method enforces that the rate of change of optical flow field is gradual along the space coordinates such as when objects in a scene undergo elastic or rigid transformations. The additional temporal smoothness constraint introduced by Toshio \textit{et al.} enforces that the acceleration field is smooth.  While these constraints may not be valid for highly dynamic scenes, they regularize the estimates and provided higher accuracy flow estimates in the presence of noise. 

     Heeger \textit{et al.} \cite{heeger1988optical} proposed a multi-frame based optical flow estimation method by sampling the frequency contents of the input frames. Heeger used 12 families of 3D spatiotemporal Gabor filters tuned to the same spatial frequency band but different temporal frequencies and spatial orientations. These filters respond to specific flow magnitudes and directions to capture optical flows. Gautama \textit{et al.} \cite{gautama2002phase} introduced a phase-based approach which measures temporal phase gradients from sequence of images to capture phase non-linearities. These phase information are then used to discard erroneous optical flow components. 

     In Kennedy \textit{et al.} \cite{kennedy2015optical}, optical flow at each time instant was estimated by fusing flow estimates from neighboring time points using a random forest classifier.  This work demonstrated the importance of incorporating positional information of occluded objects likely available in longer sequences.

     In a more recent work, Ren \textit{et al.}  \cite{ren2019fusion} (MFF) improved upon the two-frame PWC-Net \cite{sun2018pwc} method into a multi-frame method by fusing multiple flow estimates in parallel.  The MFR method  \cite{jiao2021optical} proposed a motion feature recovery approach across multiple input frames to overcome erroneous and invalid motion feature sampling in \textit{two-frame} based cost volumes mainly due to occluded and out-of-boundary scenes. MFCFlow \cite{chen2023mfcflow} extracts motion feature similarities using multiple correlation volumes and transfer these motion features across multiple frames to solve ambiguities caused by occlusion. MFCFlow achieved state-of-the-art results among all published multi-frame methods. 

     Recent state of the art methods for optical flow estimation are two-frame methods which include RAFT \cite{teed2020raft} and GMA \cite{jiang2021learning}. RAFT introduced a 4D correlation volume based on the Kronecker product of feature maps to capture visual similarities between all-pairs of pixels from two consecutive images in a sequence. RAFT also formulated the optical flow estimation problem as a recurrent problem by which convGRU blocks are used to iteratively estimate a residual flow direction from the visual similarity captured by the 4D correlation volume and context features. GMA extended the work of RAFT by introducing transformer network to capture the global dependencies of motion features in the image sequence. GMA outperformed RAFT and thus signified the importance of attention mechanism to capture long term dependencies between pixels.   

     At present, there are only a few learning based multi-frame methods for optical flow estimation with superior performance. We believe that this is mainly due to the limited availability of datasets for training multi-frame optical flow models.

\section{Model Features and Architectures}
   In this section, we present the features, design architectures, techniques and theoretical background of our model which formed the basis for our design of multi-frame optical flow estimation methods, namely SSTM and SSTM++.

\subsection{SSTM}
   Figure~\ref{fig:SSTM_arch} shows a schematic diagram of our SSTM architecture which is comprised of feature encoders and 4D correlation volumes for generating 3D motion features; and a spatiotemporal feature encoder for generating spatiotemporal context. Both the 3D motion features and spatiotemporal context information were used to update the hidden state of 3D convGRU modules and for generating an updated flow estimate.  Detailed descriptions of these units as well as the loss function used for training the SSTM model are given below.

 \begin{figure}[!ht]
        \centering
        \edef\mainLabel{fig:SSTM_arch}
        \subcaptionbox{SSTM architecture \label{\mainLabel:SSTM}}
        {
        \scalebox{2.5}{\includegraphics[width=0.4\linewidth]{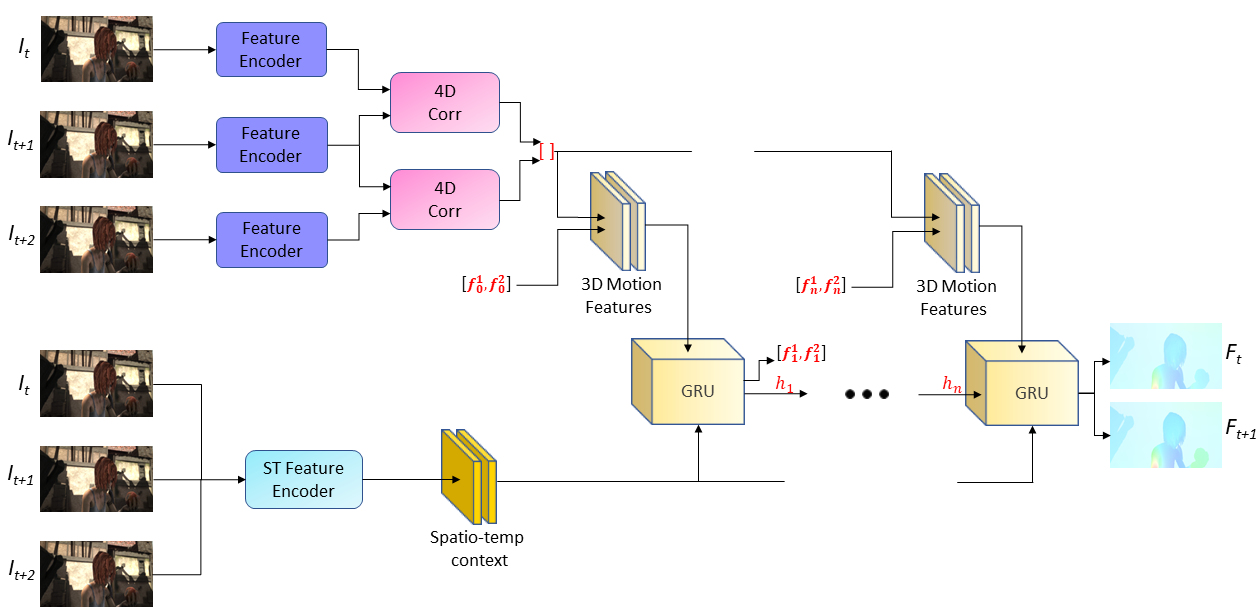}}
        }
        \subcaptionbox{3D Convolutional GRU \label{\mainLabel:convGRU}}{
        \scalebox{0.8}{\includegraphics[width=0.4\linewidth]{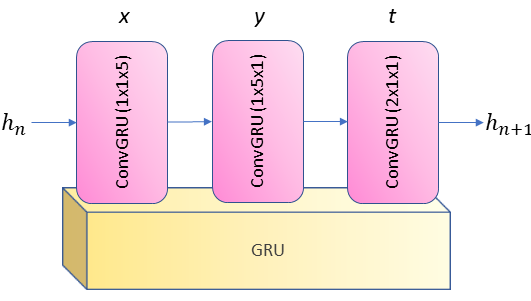}}
        }
        \subcaptionbox{Spatiotemporal context encoder \label{\mainLabel:SPT}}{
        \scalebox{1}{\includegraphics[width=0.4\linewidth]{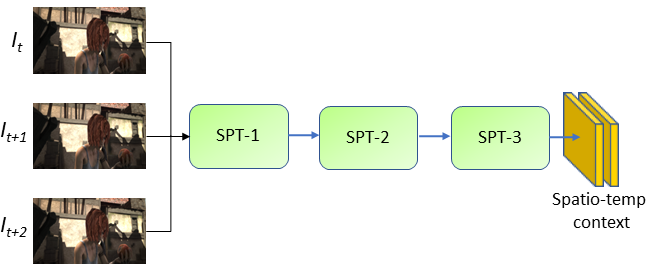}}
        }
        \decoRule
        \caption{SSTM architecture and its components. \subref{\mainLabel:SSTM} SSTM architecture comprised of 3D CNN Context feature encoder, two separate 4D correlation volumes, and 3D convGRU blocks. SSTM computes the correlation volume and context features from three input frames using the 4D correlation volume and context feature encoder. It then iteratively refines the optical flow estimate using the 3D convGRU update blocks. \subref{\mainLabel:convGRU} A 3D convGRU block built from two 1D spatial filters (along $x$ and $y$ directions) and one 1D temporal (along $t$ direction) CNN filters.
        \subref{\mainLabel:SPT} The spatiotemporal (SPT) blocks are cascaded to build the context encoder which extracts context features from three input frames.
        }
        \label{\mainLabel}
    \end{figure}

\subsubsection{Feature Encoder and 4D Correlation Volume}
    Given three input images $I_1, I_2$ and $I_3$, the feature encoder extracts spatial features of the input images at 1/8 resolution. The feature encoder was constructed using fully connected convolutional layers with differing strides such that the output feature maps are at 1/8 resolution of the input images. For an input image $I_1 \in \mathbb{R}^{C \times H \times W}$, the feature encoder gives an output feature map, $fmap1 \in \mathbb{R}^{D \times H/8 \times W/8}$, for $D = 256$ number of feature maps. We then use a 4-D correlation volume as defined in RAFT \cite{teed2020raft} to measure two multiscale visual similarities, $\mathbf{C^1}$ and $\mathbf{C^2}$ $ \in \mathbb{R}^{D \times H/8 \times W/8}$ for D=324, between pairs of neighboring feature maps in the sequence, $fmap1$ and $fmap2$, and $fmap2$ and $fmap3$, respectively. We further concatenate these two correlation volumes as $\mathbf{C} = [\mathbf{C^1}, \mathbf{C^2}]$ along a temporal dimension such that, $\mathbf{C} \in \mathbb{R}^{D \times 2 \times H/8 \times W/8}$.

\subsubsection{Spatiotemporal Context Feature Encoder}\label{sec:3.3.2}

   Spatiotemporal filters are 3D motion filters that can be applied in space-time domain to capture 2D spatial motion features (speed and direction) and 1D temporal features (interpretation and generalization of motion from the temporal queue) \cite{heeger1988optical}. Such spatiotemporal filters are widely used to learn and represent motion features in videos and sequence of images from the same scene \cite{ heeger1988optical, qiu2017learning, teney2016learning, tran2015learning, xie2018rethinking, sun2015human}. 

   Our spatiotemporal context feature encoder learns both spatial and temporal features from multiple input images. We designed such a context feature encoder using four types of spatiotemporal feature extractor blocks, namely SPT1, SPT2, SPT3 and SPT4 as shown in Figure~\ref{fig:SPTskipGRU:SPT1234}. Assuming symmetrical filter coefficients, these blocks were made of separable 3D CNN layers decoupled into separate spatial and temporal layers with residual connections. For example, a 3D convolutional filter of size $3\times3\times3$ can be decoupled into a 2D spatial filter of size $1\times3\times3$ and a 1D temporal filter of size $3\times1\times1$. Our strategy to decouple 3D convolutional layers significantly reduced the high computational demands and memory cost that we observed when using 3D convolutions. We minimized any approximation error due to our symmetric filter assumption by using multiple types of separable 3D convolutions with residual connections. 
   
   Xie \textit{et al.} \cite{xie2018rethinking} showed that such separable filters are not only computationally efficient, but also facilitate more accurate results when compared to the Inception 3D model \cite{carreira2017quo} where 3D convolutional filters of the form $t\times h\times w$ are used to learn spatiotemporal features of input videos. Similar types of blocks with separable filters are used in spatiotemporal feature representations of videos \cite{jiang2019stm, tran2018closer, yang2020spatial, li2019multi, sun2015human}.

   The context encoder consists of 6 SPT blocks that are cascaded in a specific order with differing stride values in such a way that the output feature map is at $1/8$th spatial and $2/3$rd temporal resolution as shown in Figure~\ref{fig:SSTM_arch:SPT}. Three consecutive images, $I_1, I_2$ and $I_3$, were concatenated along a temporal dimension while retaining their temporal order as $x \in \mathbb{R}^{C \times T \times H \times W}$, where $C, T, H$ and $W$ denote the number of color channels, the number of concatenated frames, height and width of each frame, respectively. For a concatenated input $x$, the cascaded SPT blocks output spatiotemporal context features, $context \in \mathbb{R}^{D \times 2 \times H/8 \times W/8}$ for $D=128$ number of 3D feature maps those capture motion features. In this work, we chose the number of frames as $T=3$.

 \begin{figure}[ht]
        \centering
        \edef\mainLabel{fig:SPTskipGRU}
        \subcaptionbox{SPT blocks \label{\mainLabel:SPT1234}}{
        \scalebox{1.8}{\includegraphics[width=0.4\linewidth]{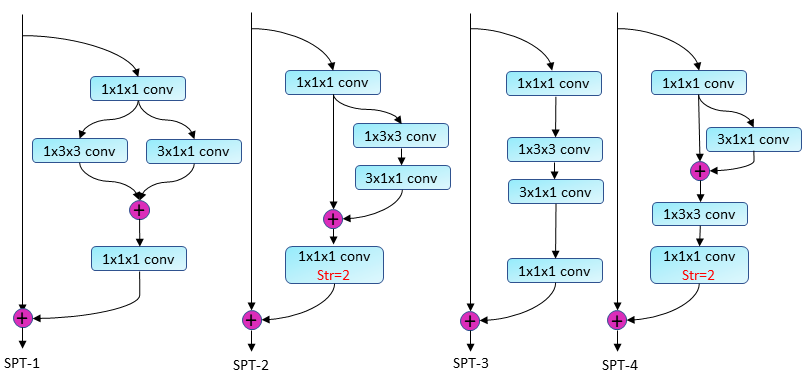}}
        }
        \subcaptionbox{Residual connections in 3D GRU \label{\mainLabel:skipGRU}}{
        \scalebox{1.8}{\includegraphics[width=0.4\linewidth]{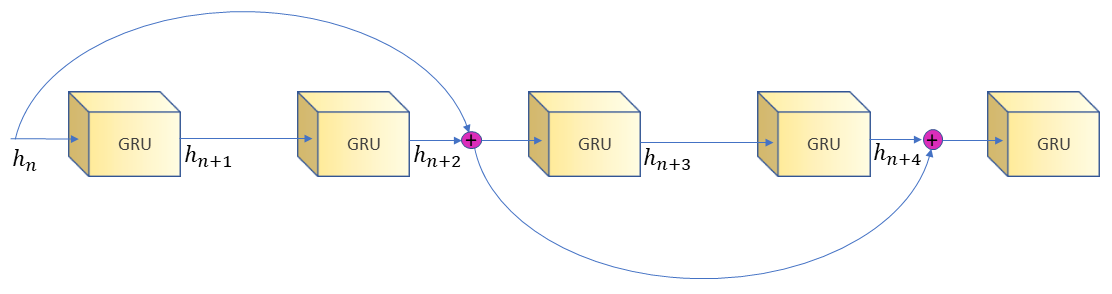}}
        }
        \decoRule
        \caption{SSTM components \subref{\mainLabel:SPT1234}: Several separable 1D temporal filters and 2D spatial filters were cascaded to build the spatiotemporal context encoders (SPT blocks).
        \subref{\mainLabel:skipGRU}: Example of residual connections of GRU hidden unit at a specified interval $r$ with $r=2$ shown for illustration.
        }
        \label{\mainLabel}
    \end{figure}

\subsubsection{3D Convolutional GRU Update Block}
   We introduce a 3D convGRU update block using a 3D spatiotemporal gating system. This update block contains $N$ GRU blocks with tied weights such that the hidden state of the $n^{th}$ 3D GRU, $h_n \in \mathbb{R}^{D \times M \times H/8 \times W/8}$, contains $D=128$ number of 3D spatiotemporal hidden units with $M$ number of motion features or optical flows estimated in parallel. In this work, we chose $M=2$ as we estimated two optical flows in parallel from three input images. We anticipate the spatiotemporal hidden units to learn recurrent dependencies in space and time among neighboring 3D locations.
   
   Input to the $n^{th}$ GRU block, $X_n$, was comprised of the correlation volume $\mathbf{C} = [\mathbf{C^1}, \mathbf{C^2}]$, previous flow estimates $f_{n-1} = [f_{n-1}^1, f_{n-1}^2]$, context features, and multi-level brightness errors $\epsilon_{n-1}^{123}$ based on the previous flow estimates.
        \begin{align}
            X_n = [\mathbf{C}, f_{n-1}, context, \epsilon_{n-1}^{123}]
        \end{align}
   The \textit{Update} and \textit{Reset} GRU equations are given as follows:
        \begin{align}
            Z_n = \sigma (Conv3d([h_{n-1}, X_n], W_z)) \\
            r_n = \sigma (Conv3d([h_{n-1}, X_n], W_r))
        \end{align}
   \noindent where, $X_n$ represents the concatenated input, $h_{n-1}$ represents the previous state and $W_z$ represents filter coefficients (weights) in the 3D convolution layer. These 3D convolutional layers are decoupled into three 1D convolutional layers corresponding to $x, y$ and $t$ coordinates. Figure~\ref{fig:SSTM_arch:convGRU} shows our 3D convGRU comprised of two 1D spatial filters and one 1D temporal filter.

   The candidate hidden state and the output vector from level $n$ were calculated as follows:
        \begin{align}
            h_n' = tanh (WX_n + r_n \odot h_{n-1})\\
            h_n = Z_n \odot h_{n-1} + (1-Z_n) \odot h_n' 
        \end{align}
   The output hidden state of the $n^{th}$ GRU, $h_n \in \mathbb{R}^{D \times M \times H/8 \times W/8}$, was split into two equal parts, namely $h_n^1$ and $h_n^2$, along the temporal dimension such that $h_n^1, h_n^2 \in \mathbb{R}^{D \times H/8 \times W/8}$.  These two hidden states $h_n^1$ and $h_n^2$ were used to generate intermediate flow updates $\Delta f_n^1$ and $\Delta f_n^2$ respectively at level $n$ using two convolutional layers with shared weights.  Both flow estimates at the $n$th level were updated as
        \begin{align}
            f_{n}^1 = f_{n-1}^1 + \Delta f_n^1 \\\
            f_{n}^2 =f_{n-1}^2  + \Delta f_n^2 \\\
            f_{n} = [f_{n}^1, f_{n}^2] 
        \end{align}

    We used $N=12$ number of such 3D convGRU blocks with tied weights during training. As shown in Figure~\ref{fig:SPTskipGRU:skipGRU}, residual connections of GRU hidden states from experimentally determined intervals were used to update the GRU inputs.  Our experimental observations suggest that these residual connections play a major role in mitigating the vanishing gradient problem as the number of GRU blocks, $N$, grows. The residual GRU connection at an interval $r$ was defined as follows:       
    \begin{equation}\label{eq:55}
           h_{n+k}=
         \begin{cases}
          GRU(X_{n+k-1}, h_{n+k-1}) + h_n &  k = r, 2r, 3r,... \\
          GRU(X_{n+k-1}, h_{n+k-1})  & \text{Otherwise} 
        \end{cases}
    \end{equation}

\subsubsection{Loss Function}

     The network generates a sequence of $N$ pairs of optical flow estimates, $\{(\bm{f}_1^1,\bm{f}_1^2), (\bm{f}_2^1,\bm{f}_2^2),..., (\bm{f}_N^1,\bm{f}_N^2)\}$ to estimate the final optical flow fields $(\bm{f}_N^1,\bm{f}_N^2)$. Intermediate flow estimates from the $i$th level $(\bm{f}_i^1,\bm{f}_i^2)$ is comprised of an intermediate flow estimate $\bm{f}_i^1$ between images $I_1$ and $I_2$; and an intermediate flow estimate $\bm{f}_i^2$ between images $I_2$ and $I_3$. Loss or objective function for supervised training was defined as the average $L_1$ norm between the estimated pair of optical flows and their ground truth values, $(\bm{f}_{gt}^1, \bm{f}_{gt}^2)$, weighted by an exponential factor $\gamma$. The following loss function was used for end-to-end training of our networks. 

    \begin{align}\label{eq:56}
        Loss_1 = \sum_{i=1}^{N} \gamma^{N-i} \frac{(\norm{\bm{f}_{gt}^1 - \bm{f}_i^1}_1 + \norm{\bm{f}_{gt}^2 - \bm{f}_i^2}_1)}{2}
     \end{align}

     For KITTI 2015, ground truth flows are available only for a single pair of images in every sequence with 20 images per sequence. To handle this limitation, we used a modified loss function while fine-tuning our models on KITTI 2015 training datasets. For any image pair $I_2$ and $I_3$ with the ground truth flow field $\bm{f}_{gt}^2$, three images $I_1$, $I_2$ and $I_3$ were used for optical flow estimation.  For fine-tuning our models on the KITTI 2015 datasets, the loss function was modified as
     \begin{align}\label{eq:57}
        Loss_2 =\sum_{i=1}^{N} \gamma^{N-i}\norm{\bm{f}_{gt}^2 - \bm{f}_i^2}_{1}
     \end{align}
   
\subsection{SSTM++}

\begin{figure}[ht]
   \includegraphics[width=1\textwidth]{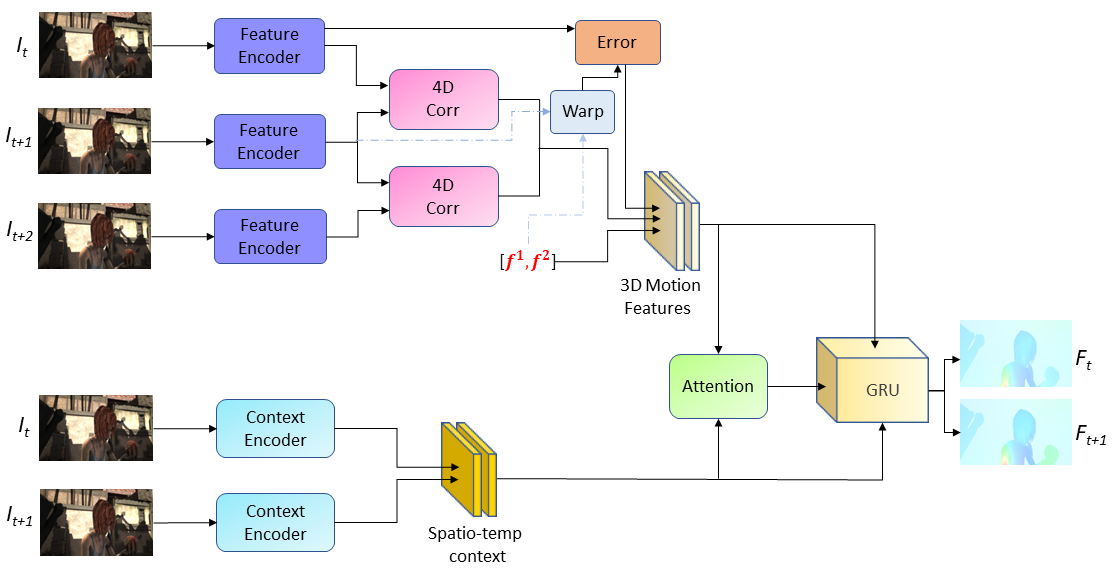}
   { \caption{SSTM++ architecture is similar to our first SSTM model with the following two additional features, namely an \textit{intermediate reconstructed-feature error} computed from feature maps warped using intermediate flow estimates and a measure of global dependency among motion features estimated using a \textit{space-time attention mechanism}. In SSTM++, we used two separate spatial context feature encoders from the first two frames using 2D convolution (to minimize the number of trainable parameters).\\}
  \label{fig:SSTM++_arch}}
\end{figure}

   Figure~\ref{fig:SSTM++_arch} shows the schematic diagram of our second method that we refer to as SSTM++.  The core architecture of the SSTM++ model is same that of SSTM.  SSTM++ has two additional modules namely a space-time attention network and a brightness error block. Our first hypothesis is that the SSTM++ method can improve over the SSTM method by using global motion features and identifying feature dependencies in both space and time using a space-time attention module. The second hypothesis is that the brightness error block can serve as an intermediate measure of confidence on intermediate optical flow fields estimated from each GRU block and therefore can improve the performance of the SSTM method. A detailed description of the attention network and brightness-based error block are given below.

   \subsubsection{Attention}
   We utilized a space-time attention mechanism to learn the global motion dependencies from a multi-frame sequence. Using multiple decompositions of attention modules over space and time, Anurag \textit{et al.} \cite{arnab2021vivit} introduced several variants of space-time transformers for video classification. Yang \textit{et al.} \cite{yang2022recurring} combined recurrent neural networks and space-time attention to efficiently achieve video action recognition. Similar to GMA \cite{jiang2021learning}, we used self attention module where the context feature is used to generate \textit{query} and \textit{key} vectors, and 3D motion features were used to generate \textit{value} vectors. Unlike GMA, however, our space-time attention design captures any global spatiotemporal dependencies and interactions among multiple image frames as well as captures any recurrent structure in the underlying fields using 3D convGRU blocks.

   As stated in Section \ref{sec:3.3.2}, the context feature, $Context \in \mathbb{R}^{D \times T \times H/8 \times W/8}$, extracted by the \textit{spatiotemporal context encoder} have a temporal length of $T=2$. This context feature is split into two spatiotemporal context features along the temporal dimension. Let these split context features be $X^1$ and $X^2$. Given the context features, $context = [X^1, X^2]$, and the 3D global motion features, $M = [M^1, M^2]$, built from the intermediate flow estimates and the correlation volume, our attention module captures global dependencies of the motion sequence in two time windows that correspond to the optical flow between $I_1$ and $I_2$ and between $I_2$ and $I_3$.

   We flattened both the context and 3D motion features along the spatial dimensions. At time $t$, the flattened context and 3D motion features were given as $X^t \in \mathbb{R}^{N\times L_c}$, and $M^t \in \mathbb{R}^{N\times L_m}$, where $N=HW$, $L_m$ and $L_c$ denote flattened channel dimensions, and $M$ represents 3D motion features.

   For projection functions $\theta(\cdot)$, $\phi(\cdot)$ and $\sigma(\cdot)$, we define the query, key and value vectors as:
    \begin{align}
      \theta(C^t) = \mathbf{W_q}C^t\\
      \phi(C^t) = \mathbf{W_k}C^t \\
      \sigma(M^t) = \mathbf{W_v}M^t
    \end{align}

   \noindent where, $W_q$, $W_k$ and $W_v$ are the query, key and key projection vectors respectively. For input frames $I_t$ and $I_{t+1}$, the aggregated global motion feature $Y^t$ is given as
   \begin{align}
       Y^t = M^t + \alpha \sum_{j=1}^Nf(\theta(C^t), \phi(C^t))\sigma(M^t)
   \end{align}
   Finally, the context $C$, global motion features ($Y$), and motion features ($M$) were concatenated as
   \begin{align}
      output = \bigr[ [C^t|Y^t|M^t], [C^{t+1}|Y^{t+1}|M^{t+1}]\bigr]
   \end{align}
   and fed as an input to the GRU block between two time instants $t$ and $t+1$.

\subsubsection{Error Block}\label{sec:3.3.3}
    We used image warping to compute multiple brightness errors at different levels. FlowNet2 \cite{ilg2017flownet} demonstrated that feeding brightness errors to multiple stacks of networks improved the accuracy of estimated optical flows. In FlowNet2, brightness error is computed as the difference between the first image in the sequence and an estimate of the second image obtained by warping the first image using the current flow estimates. In our method, however, we compute brightness error as the difference between the feature map $fmap1$ and an estimate of the feature map $fmap1$ obtained by warping $fmap2$ using the optical flow estimate from level $n$ computed as
       \begin{align}
           \hat{fmap1} &= W(fmap2;F_n^1)\\
            &=fmap2(x(t)+u(t), y(t)+v(t))
       \end{align}
    \noindent where, $fmap2$ represents feature maps extracted from the second input image; and $u(t)$ and $v(t)$ represent the components of the $n^{th}$ flow estimate of $F^1$ in the $x$ and $y$ directions respectively ($x$ and $y$ components of $F_n^1$). Brightness error $\epsilon_n^1$ at the $n$th level was computed as
    \begin{equation}
        \epsilon_n^1 = \norm{W(fmap2;F_n^1) - fmap1} 
    \end{equation}
        
    By taking advantage of the fact that our method takes multiple input images and estimates two output optical flows, we compute the following additional brightness errors at each level $n$ as:
       \begin{align}
           \epsilon_n^2 = \norm{ W(fmap3;F_n^2) - fmap2}\\\
           \epsilon_n^3 = \norm{W(W(fmap3;F_n^2);F_n^1) - fmap1} 
       \end{align}

    We feed this multi-level brightness errors $\epsilon_n^{123} = [\epsilon_n^1, \epsilon_n^2, \epsilon_n^3]$ to the GRU update block as shown in Figure~\ref{fig:SSTM++_arch}.

\section{Experiments}
\label{sec:experiments}

    Our multi-frame methods for optical flow estimation, SSTM and SSTM++ were implemented in PyTorch.  These models were trained and validated using two RTX 8000 GPUs. In this section, we present details about the procedures used for training our models, and the details of ablation experiments conducted to validate the significance of various modules and intermediate features.  In addition, we present qualitative and quantitative results of our models and compare them with other recent state of the art \textit{two-frame} and \textit{multi-frame }methods (RAFT \cite{teed2020raft}, GMA \cite{jiang2021learning}, and MFCFlow \cite{chen2023mfcflow}).

    \subsection{Datasets}
     Due to the multi-frame nature of our methods, we require the training datasets to have three or more images per sequence. The FlyingChairs \cite{dosovitskiy2015flownet} dataset is commonly used for training two-frame methods and is comprised of sequences with only two images per sequence. Therefore, for the first two stages of training our models, we used the Virtual KITTI2 \cite{cabon2020virtual} and Monkaa \cite{mayer2016large} datasets instead of the FlyingChairs dataset. The Virtual KITTI2 dataset is an updated version of Virtual KITTI \cite{gaidon2016virtual} dataset containing images from a stereo camera for each sequence. We then finetuned this result on 3DFlyingThings \cite{mayer2016large}. We further finetuned this result using Sintel \cite{butler2012naturalistic} and KITTI 2015 \cite{menze2015object} training benchmark datasets. For further finetuning and evaluation, we used the standard Sintel and KITTI 2015 training and testing benchmark datasets, respectively.
     
    \subsection{Training Schedule}
     All the models were trained in five stages and their performances were validated using the Sintel and KITTI 2015 datasets (target datasets). As in the recent optical flow estimation methods \cite{dosovitskiy2015flownet, sun2018pwc, teed2020raft, zhang2021separable, long2022detail}, we trained our models first on datasets different from the target datasets and then finetuned the models on the target datasets. 
     
     For the first three stages, we trained using the Virtual KITTI ($\mathbb{V}$), Monkaa ($\mathbb{M}$) and FlyingThings ($\mathbb{T}$) datasets which have distinctive characteristics when compared with the Sintel ($\mathbb{S}$) and KITTI 2015 ($\mathbb{K}$) target datasets. In brief, these datasets ($\mathbb{V}$, $\mathbb{M}$, $\mathbb{T}$) are larger in size and they differ in their image characteristics when compared to the target datasets ($\mathbb{S}$, $\mathbb{K}$). This helps our model to have a better generalization on unseen input images / sequences with differing features and motion patterns. In the last two stages of our training schedule, we finetuned our initial models on target datasets for final evaluation. In each of the five training stages, model performances were evaluated every 5k training iterations. A summary of these training stages is given in Table \ref{tab:table-schedule}.
     

     \begin{table}[hbt!]
       
         \caption{Summary of training schedule. $Loss_1$ and $Loss_2$ refer to loss functions defined in eq. \ref{eq:56} and eq.  \ref{eq:57}. $\mathbb{V}, \mathbb{S}, \mathbb{M}, \mathbb{T}, \mathbb{K}$ and $\mathbb{H}$ refer to Virtual KITTI2, Sintel, Monkaa, FlyingThings, KITTI2015 and HD1k datasets respectively.}
         {\scalebox{0.689}{
   \begin{tabular*}{\textwidth}{@{\extracolsep{\fill}} ll *{6}{c} }
   \cmidrule{1-8} 
   \multirow{2}{*}{\kern10pt Stage }
   &\multirow{1.5}{*}{ \kern10pt Training}
   &\multirow{1.5}{*}{Validation}
   &\multirow{1.5}{*}{ Loss}
   &\multirow{2}{*}{Input size}
   &\multirow{1.5}{*}{ Batch}
   &\multirow{1.5}{*}{ Initialized}
   &\multirow{1.5}{*}{Training samples}\\

   \multirow{2}{*}{\kern10pt }
   &\multirow{1.5}{*}{ \kern23pt data}
   &\multirow{1.5}{*}{data}
   &\multirow{1.5}{*}{ function}
   &\multirow{2}{*}{}
   &\multirow{1.5}{*}{size}
   &\multirow{1.5}{*}{ weights}
   &\multirow{1.5}{*}{(triplets)}\\
   
   \cmidrule{1-8}

   \kern10pt \textit{vkitti}  & \kern30pt $\mathbb{V}$ & $\mathbb{V}$ &  $Loss_1$ & [288, 960] & 8 & - & 38,035\\ 
   \kern10pt \textit{monkaa} & \kern30pt $\mathbb{M}$ & $\mathbb{S}$ & $Loss_1$ & [456, 720]  & 6 & \textit{vkitti} &34,384\\
   \kern10pt \textit{things} & \kern30pt $\mathbb{T}$ & $\mathbb{S}$ & $Loss_1$ & [400, 720]  & 6 & \textit{monkaa} &44,780\\
   \kern10pt \textit{sintel} & \kern5pt $\mathbb{V+T+S+H}$  &  $\mathbb{S}$  & $Loss_1$ & [368, 768] & 6 & \textit{things} & 147,089\\ 
   \kern10pt \textit{kitti} & \kern30pt $\mathbb{K}$ &  $\mathbb{K}$  & $Loss_2$ & [288, 960] & 6 & \textit{sintel} & 200\\ 

   \cmidrule{1-8}
   \end{tabular*}}}
         \label{tab:table-schedule} 
    \end{table}

     \subsubsection{Stage \romannumeral 1: vkitti}
     In the first training stage \textit{vkitti}, we used the Virtual KITTI2 datasets (dataset effectively used at this stage abbreviated as $\mathbb{V}$) for both training and validation. We split the Virtual KITTI2 dataset into two non-overlapping sets of training (90\%) and validation (10\%) datasets. The models were trained using 38,035 triplet sequences (i.e., three consecutive frames or images per sequence) of $288 \times 960$ spatial pixel resolution with a batch size of 8 and a learning rate of $4 \times 10^{-4}$ for 100k iterations using the loss function given in eq. \ref{eq:56}. At this training stage, the learning rate is relatively higher than that of the subsequent training stages.

     \subsubsection{Stage \romannumeral 2: monkaa}
     In the second training stage \textit{monkaa}, our models were initialized with the network weights from Stage \romannumeral 1 \space \textit{vkitti} and finetuned / trained using the Monkaa dataset and validated using the Sintel training dataset (dataset effectively used at this stage abbreviated as $\mathbb{V+M}$).  The Monkaa training dataset was comprised of 34,384 triplet sequences of $456 \times 720$ spatial pixel resolution. Because each of the image triplets in this set has two consecutive ground truth optical flow values, we used the loss function described in eq. \ref{eq:56}. We trained this stage for 100k iterations with a batch size of 6 and a learning rate of $4 \times 10^{-4}$.

    \subsubsection{Stage \romannumeral 3: things}
     In the third training stage \textit{things}, our models were initialized with the network weights from Stage \romannumeral 2 \space \textit{monkaa} and finetuned / trained using the FlyingThings dataset (dataset effectively used at this stage abbreviated as $\mathbb{V+M+T}$). The FlyingThings training dataset was comprised of 44,780 triplet sequences of $400 \times 720$ spatial pixel resolution. All training parameters such as the batch size, loss function, number of training iterations and learning rate were same as those of Stage \romannumeral 2 \space \textit{monkaa}.  
     
     The trained model from Stage \romannumeral 3 \space \textit{things} was tested using the Sintel and KITTI 2015 \textit{training} datasets that were not seen by the model during training ($\mathbb{V+M+T}$ datasets only were used for training). The performance of our models on Sintel and KITTI 2015 training datasets were compared with other methods as presented in Table \ref{tab:table-result} (under ``Sintel (train)'').

     \subsubsection{Stage \romannumeral 4: sintel}
     In the fourth training stage \textit{sintel}, our models were initialized with the network weights from Stage \romannumeral 3 \space \textit{things} and finetuned / trained using a mixture of datasets from Sintel training, Virtual KITTI2 validation, FlyingThings and HD1K \cite{kondermann2016hci} datasets with a mixture proportion of 0.13, 0.11, 0.74 and 0.02 respectively (dataset effectively used at this stage abbreviated as $\mathbb{V+T+S+H}$). As all these benchmark datasets have ground truth for multiple input images of the same scene, we used the loss function as in eq. \ref{eq:56} for training the models. In this stage, we decreased the batch size to 6, decreased the learning rate to $1.25 \times 10^{-4}$ and trained our networks for 120k iterations. The sintel-finetuned models were evaluated using the testing datasets provided by the Sintel leaderboard. It should be noted that ground truths for these testing datasets are not publicly available. Testing results for the Sintel test dataset as reported in the \href{http://sintel.is.tue.mpg.de/results}{Sintel Leaderboard} are presented in Table \ref{tab:table-result} (under ``Sintel (test)'').

     \subsubsection{Stage \romannumeral 5: kitti}
     In the fifth (last) training stage \textit{kitti}, we initialized our models with the network weights from Stage \romannumeral 3 \space \textit{things} and finetuned / trained using KITTI2015 \textit{training} dataset. When compared to Stage \romannumeral 3 \space \textit{things}, we decreased the batch size to 6, decreased the learning rate to $1 \times 10^{-4}$ and trained our models for 70k iterations. While the KITTI training dataset has 20 consecutive images in each sequence, ground truth flows are available for only one pair of consecutive images. Therefore, we used a modified loss function as in eq. \ref{eq:57} that computes the training loss using only one of the two flow estimates and one ground truth flow available per sequence. The kitti-finetuned models were evaluated using testing datasets provided by the KITTI leaderboard for which ground truths are not publicly available.  Testing results for KITTI test datasets as reported in the \href{https://www.cvlibs.net/datasets/kitti/eval_scene_flow.php?benchmark=flow}{KITTI Leaderboard} are presented in Table \ref{tab:table-result} (under ``KITTI-15 (test)''). 

\subsection{Training and Testing Results}
     We evaluated the performance of our multi-frame optical flow estimation models on the current benchmark datasets using the standard optical flow evaluation metrics. We used Sintel and KITTI 2015 training and testing datasets to evaluate our methods and reported the endpoint error (EPE) and a measure of outlying estimates as a function of EPE (Fl-all) evaluation results. For Sintel training dataset, we reported the average EPE on both clean and final datasets at different training stages. For KITTI 2015 training dataset, we reported both the EPE and Fl-all errors. We also reported the EPE results on the testing data evaluated by the Sintel leaderboard and both the Fl-all and Fl-fg results reported by KITTI leaderboard. These results were compared with the most recent published state-of-the-art methods for optical flow estimation. In Table \ref{tab:table-result}, we present these quantitative evaluation results on Sintel and KITTI test benchmark datasets from Sintel leaderboard and KITTI leaderboard respectively.

     \begin{table}[hbt!]
       \caption{Performance of optical flow models on Sintel and KITTI 2015 benchmark datasets. '$\mathbb{C}$/($\mathbb{V+M}$)+$\mathbb{T}$' refers to methods trained on FlyingChairs or Virtual KITTI2 and Monkaa then finetuned on Flying Things.  $\mathbb{C+T+S}$/$\mathbb{K}$ ($\mathbb{+H}$) refers to using specific dataset during finetuning the model on Sintel using a mixture of datasets including Sintel, Virtual KITTI, Monkaa, Flying Things and HD1K. The models were finetuned on KITTI using the KITTI dataset only. Methods with \textcolor{red}{*} refer to multi-frame methods. Test results written in bold are the best results among the listed methods and the ones underlined are second to the best. Results with $\textbf{\dag}$ refer to warm-start flow initialization as defined in RAFT \cite{teed2020raft}. ``Ours (SSTM)'' refers to our SSTM model shown in Figure~ \ref{fig:SSTM_arch} and ``Ours (SSTM++)'' refers to our SSTM++ method shown in Figure~\ref{fig:SSTM++_arch}.}
       {\scalebox{0.67}{
\begin{tabular*}{\textwidth}{@{\extracolsep{\fill}} ll *{8}{c} }
\cmidrule{1-10} 
 \multirow{2}{*}{\kern6pt Training\kern1pt}
 &\multirow{2}{*}{\kern5pt Method}
   &\multicolumn{2}{c}{ Sintel (train)} 
   & \multicolumn{2}{c}{KITTI-15 (train)} 
   & \multicolumn{2}{c}{ Sintel (test)} 
   & \multicolumn{2}{c}{ KITTI-15 (test)} \\
 \cmidrule{3-4} \cmidrule(lr){5-6} \cmidrule(lr){7-8} \cmidrule(lr){9-10}

  \kern15pt data & & Clean & Final & APE & Fl-all(\%) & Clean & Final & Fl-fg (\%) & Fl-all (\%)  \\

\cmidrule{1-10}

    & PWC-net \cite{sun2018pwc} &  2.55  & 3.93 & 10.35 &  33.6 &  -  &  -  & \kern20pt - & -  \\
    & FlowNet2 \cite{ilg2017flownet} &  2.02  &  3.54  &  10.08 &  30.0 &  -  &  -  & \kern20pt - & -  \\
    & VCN \cite{yang2019volumetric} &  2.21 &  3.68  &  8.36 &  25.1 &  -  &  -  & \kern20pt - & -  \\
     \kern10pt $\mathbb{C}$/($\mathbb{V+M}$)+ & MaskFlowNet \cite{zhao2020maskflownet} &  2.25  &  3.61  &  - & 23.14 &  -  &  -  & \kern20pt - & -  \\
    \kern5pt $\mathbb{T}$/Autoflow & RAFT \cite{teed2020raft}& 1.43  & 2.71  &  5.04 &  17.4 &  -  &  -  &  \kern20pt - & -  \\ 
     & GMA \cite{jiang2021learning} & 1.30  &  2.74  &  4.69 & 17.1 &  -  &  -  & \kern20pt  - & -  \\ 
     
     & CRAFT \cite{sui2022craft} &  \textbf{1.27}  &  2.79  &  4.88 & 17.5 &  -  &  -  & \kern20pt  - & -  \\ 
     \cmidrule{2-10}
    & Ours(SSTM)\textbf{\textcolor{red}{*}} & 1.45 & \textbf{2.55} & 4.40 & 15.9 & -  &  -  &  \kern20pt - & -  \\
    & Ours(SSTM++)\textbf{\textcolor{red}{*}} & 1.47 &  2.71 & \textbf{4.13} &  \textbf{15.3} &  -  &  - & \kern20pt - &  - \\ 
 
\cmidrule{1-10}
    & MFF\textbf{\textcolor{red}{*}} \cite{ren2019fusion} &  -  & - & - &  - &  3.42  &  4.57  & \kern20pt 7.25 &  7.17 \\
    & PWC-Net+ \cite{sun2019models} &  1.71  & 2.34 & 1.50 &  5.30 &  3.45  & 4.60  & \kern20pt 7.88 & 7.72  \\
    
    & FlowNet2 \cite{ilg2017flownet} & 1.45  & 2.01  &  2.30 &  6.80 &  4.16  &  4.74  &  \kern20pt  - & 11.48  \\ 

    & MaskFlowNet \cite{zhao2020maskflownet} &  -  &  -  &  - & 2.52 &  4.17  &  7.70  & \kern20pt 7.70 & 6.11 \\

    & VCN \cite{yang2019volumetric} &  1.66 & 2.24 & 1.16 & 4.1 &  2.81 &  4.40 & \kern20pt 8.66 & 6.30 \\
   
    \kern10pt $\mathbb{C+T+S}$/$\mathbb{K}$  & RAFT \cite{teed2020raft} &  0.76  &  1.22  &  0.63 &  1.50  & 1.94 &  3.18&  \kern20pt 6.87 & 5.10   \\

    \kern20pt ($\mathbb{+H}$)& &  &  &   &  &  \kern3pt$1.61^\textbf{\dag}$ & \kern4pt$2.86^\textbf{\dag}$ & \kern20pt  &   \\

    & MFR\textbf{\textcolor{red}{*}} \cite{jiao2021optical} & 0.64 & 1.04 & 0.54 & 1.10 &  1.55 & 2.80 & \kern20pt - & 5.03 \\

    & MFCFlow\textbf{\textcolor{red}{*}} \cite{chen2023mfcflow} & 0.56 & 0.89 & 0.55 & 1.10 &  \underline{1.49} & 2.58 & \kern20pt - & 5.00 \\

    & GMA \cite{jiang2021learning} & 0.62  &  1.06  & 0.57 &  1.20  &  \kern5pt $\textbf{1.39}^\textbf{\dag}$  & \kern5pt $\textbf{2.47}^\textbf{\dag}$ & \kern20pt 7.03 & 5.15  \\ 
\cmidrule{2-10} 
    & Ours(SSTM)\textbf{\textcolor{red}{*}}& 0.67 & 1.06 & 0.60 & 1.37  &\kern-3.5pt 1.78 & \kern-3.5pt 3.06 & \kern20pt \textbf{6.40} &  \textbf{4.72}\\ 
        
    & Ours(SSTM++)\textbf{\textcolor{red}{*}} & 0.63 & 0.96 & 0.58 &  1.29  &\kern-3.5pt1.61  & \kern-3.5pt \underline{2.53} & \kern20pt \underline{6.71} & \underline{4.83}\\ 
    & &  &  &   &  &\kern2pt $1.59^\textbf{\dag}$  & \kern2pt $2.54^\textbf{\dag}$ & \kern10pt  &   \\

\cmidrule{1-10} 
\end{tabular*} 
}}
       \label{tab:table-result}
      \end{table}

\begin{table}[ht]
       \centering
       \setlength\tabcolsep{0pt}
       \caption{Sintel leaderboard results on Sintel-final test benchmark dataset. $d_{i-j}$ represents EPE near occluded boundaries with a distance ranging $i$ to $j$ pixels. $s_{m-n}$ represents EPE over regions with velocities between $m$ to $n$ pixels per frame. Results written in bold are the best results among the listed methods and the ones underlined are second to the best. Results with $\textbf{\dag}$ refer to warm-start flow initialization as defined in RAFT \cite{teed2020raft}. Methods with \textcolor{red}{*} refer to multi-frame methods.}
       {\begin{tabular*}{\textwidth}{@{\extracolsep{\fill}} ll *{7}{c} }
  \cmidrule{1-9} 
  \multirow{2}{*}{\kern20pt Method }
  &\multirow{2}{*}{$d_{0-10}$}
  &\multirow{2}{*}{$d_{10-60}$}
  &\multirow{2}{*}{$d_{60-140}$}
  &\multirow{2}{*}{$s_{0-10}$}
  &\multirow{2}{*}{$s_{10-40}$}
  &\multirow{2}{*}{$s_{40++}$}\\
  &  &  & & &  &  & & \\
  
 \cmidrule{1-9}

\kern7pt MFF\textbf{\textcolor{red}{*}}  \cite{ren2019fusion}  & 4.664	& 2.017 &1.222 & 0.893 & 2.902 & 26.810 \\ 

\kern7pt RAFT\cite{teed2020raft} & $3.112^\textbf{\dag}$ & \kern5pt $1.133^\textbf{\dag}$ & \kern5pt $0.770^\textbf{\dag}$ & \kern5pt $0.634^\textbf{\dag}$ & \kern5pt $1.823^\textbf{\dag}$ & \kern5pt $16.371^\textbf{\dag}$\\

\kern7pt MFR\textbf{\textcolor{red}{*}}  \cite{jiao2021optical} & 3.075 & 1.112 & 0.772 & 0.674 & 1.829 &	15.703\\

\kern7pt MFCFlow\textbf{\textcolor{red}{*}}  \cite{chen2023mfcflow} & 3.018 & 1.113 & 0.662 & 0.587 & 1.678 & 14.647\\

\kern7pt GMA \cite{jiang2021learning} & $\textbf{2.863}^\textbf{\dag}$ & \kern5pt $1.057^\textbf{\dag}$ & \kern5pt $0.653^\textbf{\dag}$ & \kern5pt $0.566^\textbf{\dag}$ & \kern5pt $1.817^\textbf{\dag}$ & \kern8pt $\textbf{13.492}^\textbf{\dag}$\\

 \kern1pt Ours(SSTM++)\textbf{\textcolor{red}{*}}  & \underline{2.899} & \kern5pt \textbf{0.929} & \underline{0.583} & \kern4pt\textbf{0.519} & \underline{1.666} & \underline{14.560} \\ 

& $2.982^\textbf{\dag}$ & \kern6pt $\underline{0.957}^\textbf{\dag}$ & \kern5pt $\textbf{0.549}^\textbf{\dag}$& \kern6pt $\underline{0.531}^\textbf{\dag}$ & \kern5pt $\textbf{1.658}^\textbf{\dag}$& 
 \kern5pt $14.573^\textbf{\dag}$\\ 

\cmidrule{1-9}
\end{tabular*} 
}
       \label{tab:table-sintel-result} 
\end{table}

\begin{figure}[ht]
\includegraphics[width=1\textwidth]{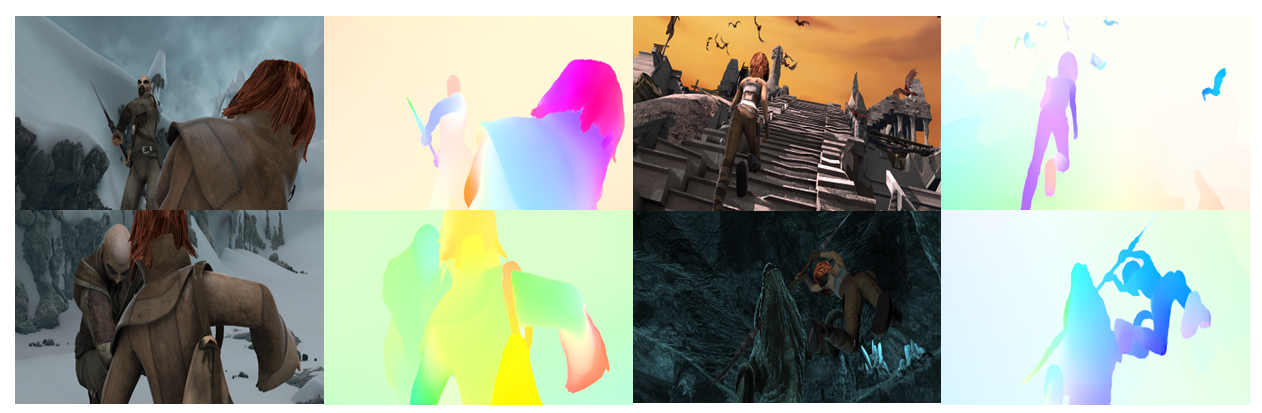}{ 
         \caption{Sample frames from the Sintel test dataset and their corresponding optical flows estimated using our SSTM++ method.\\}
         \label{fig:sintelResult}}
\end{figure}

\subsection{Results on Sintel}
          
     Regional model errors as a function of proximity to the nearest occluded regions $d_{i-j}$ and as a function of the ground truth flow magnitudes $s_{m-n}$ assessed using the EPE metric are presented in Table \ref{tab:table-sintel-result}. ``$d_{i-j}$'' represents pixel locations that are at least $i$ pixels and at most $j$ pixels away from their respective nearest occluded region.  ``$s_{m-n}$'' represents pixel locations with a minimum flow velocity of $m$ pixels/frame and a maximum flow velocity of $n$ pixels/frame.  
     
     In locations that are 0 to 60 ($d_{0-10}$) pixels away from the nearest occlusion boundaries, SSTM++ performance was comparable with the top performing method GMA \cite{jiang2021learning} (only 1.24\% higher). In other locations that are are 10 to 60 pixels ($d_{10-60}$) and 60 to 140 ($d_{60-140}$) pixels away from the nearest occluded regions, our methods significantly outperformed GMA by 12.1\% and 14.4\% respectively. 
     
     In regions with velocities between 0 to 10 ($s_{0-10}$) and 10 to 40 ($s_{10-40}$) pixels per frame, SSTM++ outperformed GMA by 10.9\% and 8.3\%, respectively. In locations with larger flow velocities ($> 40$ pixels per frame, $s_{40++}$), however, our method ranked second among all the methods that were examined with an error of 7.33\% higher than that of GMA. 

     These results strongly support our hypothesis that when compared with two-frame methods our multi-frame approach has a better generalization near occluded regions. 
     
\subsection{Results on KITTI2015}

     Table \ref{tab:table-kitti-result} shows Fl-all, Fl-fg and Fl-bg error metrics of all top performing methods in both non-occluded regions and all regions.  These errors measure the percentage of outlying estimates in the foreground, background and all regions respectively.  Outlying estimates were those with an EPE $\geq 3$ pixels of flow and and EPE $\geq 5\%$ of the magnitude of the ground truth flow vectors. Our SSTM and SSTM++ methods ranked first or second with either RAFT or GMA in the third position.  For example, Fl-all errors for our methods were 7.5 \% and 6.5 \% lower than the third position in all regions and in non-occluded regions respectively. Moreover, both SSTM and SSTM++ methods outperformed the \textit{two-frame} and \textit{multi-frame} methods in all comparison metrics (both non-occluded and all regions). These results support our hypothesis that both occluded as well as non-occluded regions can be better understood with multi-frame based optical flow estimation methods. 

  \begin{table}[hbt!]
      \setlength\tabcolsep{0pt}
      \caption{KITTI2015 leaderboard results on KITTI2015 test benchmark dataset. ``non-occ'' refers to non-occluded regions and ``all'' refers to all pixels. Fl-bg\%, Fl-fg\% and Fl-all\% refer to percentage outliers in the background, foreground and all regions, respectively. Results written in bold are the best results among the listed methods and the ones underlined are second to the best.}
      {

\begin{tabular*}{\textwidth}{@{\extracolsep{\fill}} ll *{5}{c} }
\cmidrule{1-7} 
 \multirow{2}{*}{\kern15pt Method }
 &\multirow{1}{*}{\kern1pt Fl-bg \%}
 &\multirow{1}{*}{Fl-fg \%}
 &\multirow{1}{*}{Fl-all \%}
 &\multirow{1}{*}{Fl-bg \%}
 &\multirow{1}{*}{Fl-fg \%}
 &\multirow{1}{*}{Fl-all \%}
 \\
 & (non-occ) & (non-occ) & (non-occ) & (all) & (all)  & (all) \\ 
 
\cmidrule{1-7}

\kern7pt MFF \cite{ren2019fusion}  & 4.52 & 4.25 & 4.47  & 7.15 & 7.25 & 7.17 \\

\kern7pt RAFT\cite{teed2020raft} & 2.87 & 3.98 & 3.07 & 4.74 & 6.87 & 5.10 \\ 
\kern7pt GMA\cite{jiang2021learning} & 2.97 & 3.80 & 3.12 & 4.78 & 7.03 & 5.15 \\ 

\kern1pt Ours(SSTM) &\kern-1pt\underline{2.75} & \textbf{3.43} & \textbf{2.87} & \textbf{4.39} & \textbf{6.40} & \textbf{4.72} \\ 


\kern1pt Ours(SSTM++) &\kern-1pt\textbf{2.71} & \underline{3.66} & \underline{2.89} & \underline{4.45} & \underline{6.71} & \underline{4.83} \\ 

\cmidrule{1-7}
\end{tabular*}   }
      \label{tab:table-kitti-result} 
  \end{table}

  \begin{figure}[ht]
    \includegraphics[width=1\textwidth]{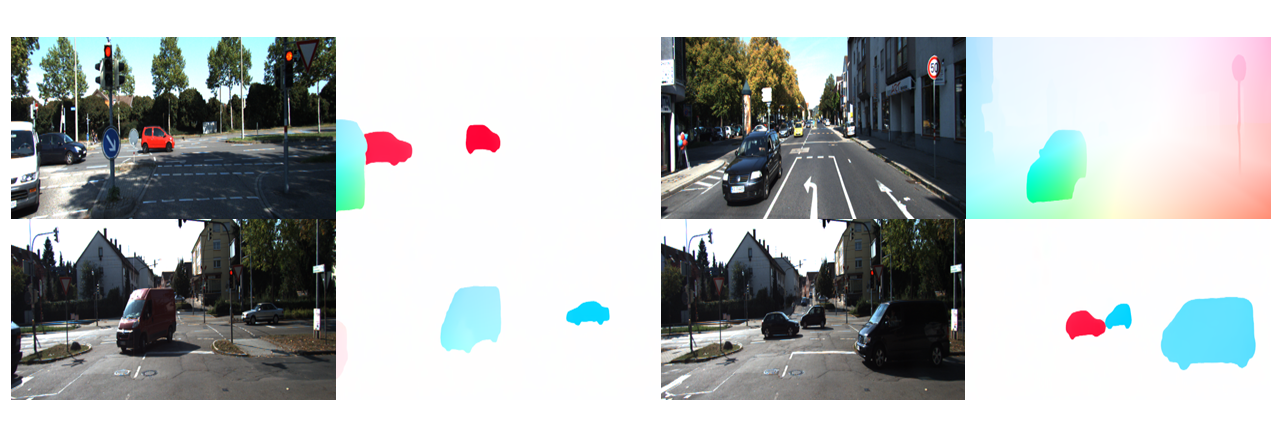}{ 
     \caption{Sample frames from KITTI2015 test dataset and their corresponding optical flow estimates using our SSTM method.}
         \label{fig:kittiresult}}
    \end{figure}

\subsection{Results on sample unseen frames}
    We further evaluated our multi-frame methods on image frames which were not part of training or validation datasets (unseen frames). Our goal is to demonstrate the generalizability of our multi-frame method in comparison with other recent state-of-the-art methods. For this experiment, we used the DAVIS dataset \cite{pont20172017} which consists of high resolution real world video frames. 
    
\begin{figure}[ht]
\includegraphics[width=1\textwidth]{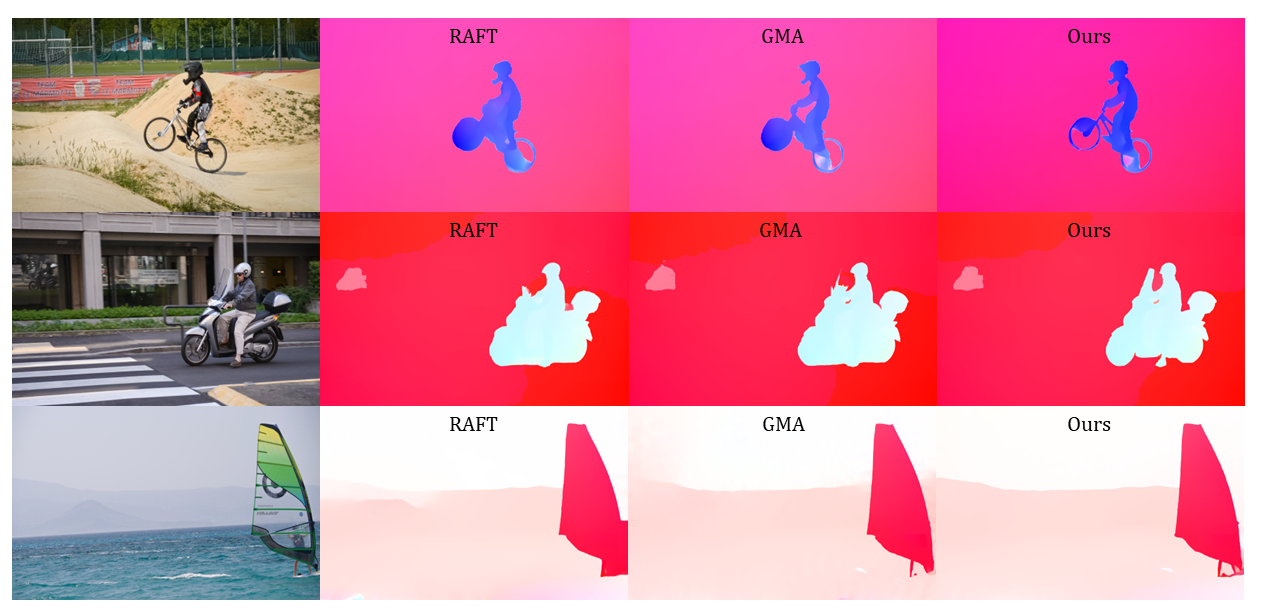}{ 
     \caption{Optical flow estimates (OFEs) of sample frames from the DAVIS dataset \cite{pont20172017} using model weights from the \textit{sintel} stage for SSTM++, RAFT and GMA. In the first row, detailed motion features and flows were distinctly better captured (the bicycle frame, the tires and the rider) by SSTM++ when compared to both RAFT and GMA. In the second row, SSTM++ captured complex moving features (transparent windshield) and boundaries (the bottom frame of the motorbike and the feet of the biker) while RAFT and GMA failed to do so. The third row demonstrates \textit{out-of-boundary} motion where the surfboard and the sail partially leave the scene in the next frame. SSTM++ was able to learn the motion features and flow patterns from the previous scene to better estimate the optical flow compared to RAFT and GMA (legs of the surfer and surfboard were accurately captured by SSTM++).\\}
    \label{fig:davisResultComp}}
\end{figure}

    In Figure~\ref{fig:davisResultComp}, we compared optical flow estimates of our SSTM++ method on selected sample frames from the DAVIS dataset with those of RAFT and GMA. In all the three methods, we used the model from Sintel test evaluation (i.e., weights from the \textit{sintel} stage). It is evident that detailed motion features and flow estimates in complex regions (with transparent objects and out-of-boundary motions) were distinctly better than the competing methods.

\subsection{Ablation Experiments}
    To determine the significance of each of the main parts of our design, we conducted ablation experiments by training different versions of our methods on Virtual KITTI and Monkaa datasets and evaluated the result on Sintel (Clean and Final) and KITTI2015 training datasets. In brief, various ablated models were created to assess the significance of using 1) 3D convolution in place of 2D convolution while building the context features, 2) attention features, and 3) feature warping for estimating intermediate optical flow estimation errors. The list of features used to build different versions of our methods is shown in Table \ref{tab:table-abl-result}.

    We first trained each of these variants of our methods for 50k iterations with batch size of 8. We then finetuned this result on Monkaa dataset with a batch size of 6 while validating on Sintel training dataset. The validation result on Sintel dataset is then used to compare these features and their significance on the final model. Moreover, we also compared the total number of learnable parameters with or without the ablated features listed in Table \ref{tab:table-abl-result}. These ablation experiments are conducted using two RTX 8000 GPUs. Quantitative performance of the models with and without ablated features is presented in Table \ref{tab:table-abl-result}. A detailed description of the ablated features are presented in the following subsections.

    \subsubsection{Context}
    We designed two types of spatiotemporal encoders to capture the context information from the image sequences. The first type of encoder utilized 2D convolution to capture the spatial features of the first two images $I_1$ and $I_2$ in the sequence. These images pass through 2D CNN blocks with shared weights. The output spatial feature maps are then concatenated along an additional temporal dimension to form a 3D spatiotemporal features. In the second context encoder design, on the other hand, we used a 3D CNN based architecture as described in Figure~\ref{fig:SSTM_arch:SPT} to capture the spatiotemporal context features.  In this configuration, all the three images were concatenated and fed to the cascaded SPT blocks.

    In our ablation experiment, we used a 3D CNN context encoder in our SSTM method and a 2D CNN context feature encoder in our SSTM++ method. By using our separable 3D CNN based spatio-temporal context encoder, we were able to get comparable results with the 2D CNN based context encoder while reducing the learnable parameters by about half a million. However, the 3D CNN based method had a higher GPU memory usage during training due to the cubic order of the feature maps size.


\subsubsection{Attention}
    We assessed two models to analyze the effect of our space-time attention module on the number of learnable parameters, inference time and standard results on Sintel and KITTI2015 training datasets. In the first model, we used a spatiotemporal attention module and a 2D CNN context encoder. In the second model, we used a 3D CNN context encoder without any attention module. 

    The main difference between SSTM and SSTM++ methods is the use of space-time attention mechanism only in SSTM++. Our initial hypothesis is that by introducing space-time attention networks to our SSTM method, we can have a better understanding of the global dependencies of the input sequences in space-time domain and have a more accurate and generalized optical flow estimation, especially in/near occluded regions. Our results on Sintel test benchmark dataset supports this hypothesis. Results in Table \ref{tab:table-result} show that SSTM++ improved the EPE results approximately by 10 \% and 17\% on Sintel clean and Sintel final test dataset, respectively. More detailed results in Table \ref{tab:table-sintel-result} show that SSTM++ outperforms SSTM in all listed regions around occluded boundaries and velocities. However, we did not observe a similar result improvement in KITTI test dataset. In fact, SSTM performs better than SSTM++ on KITTI. Similar results were also reported by others \cite{teed2020raft, jiang2021learning}. We believe that the shorter sequence lengths in KITTI is primarily making it harder for SSTM++ to understand longer space-time dependencies in each scene (i.e. difficulty in associating similar features that are separated by larger extents in space and/or time).

\subsubsection{Warping}
    The error block defined in Section \ref{sec:3.3.3} generates a brightness error map by warping features using multi-level optical flow estimates. In our ablation experiment, we compared the effect of feeding a multi-level brightness error to the 3D motion extractor block.

    Results in Table \ref{tab:table-abl-result} show that multi-level brightness error maps improved optical flow estimates in Sintel clean and Sintel final validation sets . This observation demonstrates that such warping based intermediate error maps can improve optical flow estimates.  

\begin{table}[!ht]
    \caption{ Ablation experiment results. Trained on $\mathbb{V+M}$ (80k on Virtual KITTI and 50k on Monkaa) and evaluated on Sintel and KITTI 2015 training datasets. Inference time is measured using Sintel sequences (clean and final). Underlined features are used in the final SSTM++ (with attn) method, and features with '*' are used in the final SSTM (without attn) method.}

 {\scalebox{0.67}{
\begin{tabular*}{\textwidth}{@{\extracolsep{\fill}} ll *{7}{c} }
   \cmidrule{1-9} 
   \multirow{2.5}{*}{\kern6pt Experiment }
   &\multirow{2.5}{*}{\kern16pt Feature} &\multicolumn{2}{c}{ Sintel (train)} & \multicolumn{2}{c}{ KITTI-15 (train)} &\multirow{2.5}{*}{Parameters (M) \kern6pt} &\multirow{2.5}{*}{Inference time (s)\kern6pt}\\
   \cmidrule{3-4} \cmidrule(lr){5-6} 
   & & Clean & Final & AEPE & Fl-all(\%) &  \\

   \cmidrule{1-9}
   &\underline{2D Convolution} &  2.17  &  2.94  & 5.92 & 18 &  5.61  &  0.38 & \\
   \kern7pt Context  & 3D Convolution* & 1.88  &  2.81  &  5.58 &  19  &  5.12 &  0.34 \\ 
  \cmidrule{1-9}
   & No attention* &  1.67  &  2.95  & 5.05 & 17.3& 5.12  & 0.41 & \\
   \kern7pt Attention & \underline{Spatio-temp} & 1.60& 3.00  & 4.35 & 15.6 &  6.09  &  0.48 &\\
   \cmidrule{1-9}
   & No warping* &  2.56  &  3.19  & 5.32 & 17.1 & 5.89 & 0.32 & \\
   \kern7pt Warping & \underline{Warping} & 2.17  &  2.94  &  5.92 & 18 & 6.09 & 0.38 &  \\  
   \cmidrule{1-9}
\end{tabular*}
}}
\label{tab:table-abl-result}
\end{table}

\subsection{Effect of Warm-start}

    RAFT \cite{teed2020raft} introduced a warm-start procedure to initialize the optical flow estimates using previously estimated optical flows from the neighboring frames. In our previous work \cite{ferede2022multi}, we analyzed the effect of such initialization on our multi-frame methods and we observed that our methods did not benefit from two-step warm-start initialization. In brief, given two flow estimates, $\bm{f}_1$ and $\bm{f}_2$ from the first three input images, $I_1, I_2$ and $\bm{f}_3$, the next flow estimates $\bm{f}_3$ and $\bm{f}_4$ are initialized by $\bm{f}_1$ and $\bm{f}_2$ respectively. We believe that this is mainly due to a wider temporal gap (two time-steps) between the time-step at which the previous flow was estimated and the time-step that required flow initialization. Given the higher flow velocities and non-rigid motions in the Sintel dataset, the scenes could undergo significant changes within two temporal steps causing the initialization to be erroneous.

    In this work, we experimented a different approach by which we only initialized the first of each pair of flow estimates with the second flow from the previous pair of estimates. In other words, given two flow estimates, $\bm{f}_1$ and $\bm{f}_2$ from the first three input images, the next flow estimates $\bm{f}_3$ and $\bm{f}_4$ are initialized as $\bm{f}_3=\bm{f}_2$ and $\bm{f}_4=\bm{0}$ respectively. With this initialization approach, we were able to observe a slight improvement on Sintel Clean test dataset. 

    Nonetheless, as shown in Table \ref{tab:table-result}, it can be observed that the performances of RAFT and GMA are highly affected by the use of warm-start initialization. More specifically, these methods significantly underperformed without warm-start initialization. In contrast, SSTM++ was able to outperform these methods when warm-start initialization is not used. 
\section{Software}
SSTM models along with necessary instructions for running the software will be available to the public in the following URL:\\ \href{https://github.com/Computational-Ocularscience/SSTM}{https://github.com/Computational-Ocularscience/SSTM}.

\section{Conclusion}
    In this work, we introduced two multi-frame optical flow estimation methods (SSTM, SSTM++) on the basis of understanding the spatiotemporal nature and dependency of the input sequence across multiple temporal cues. Our multi-frame approach utilizes recurrent transformers that can capture the global spatial and temporal feature dependencies as well as the space-time recurrent nature of the input frames. These salient features of our methods allow them to better understand ambiguities caused by occlusion and out-of-boundary regions (mainly due to lack of local evidences) and provide accurate flow estimates. Based on qualitative and quantitative assessment using standard benchmark optical flow datasets (both real world and synthetic), our methods achieved high performance in/near occluded and out-of-boundary regions compared to recent state-of-the-art multi-frame and two-frame methods. Moreover, we achieved state-of-the-art results in KITTI2015 and Sintel Final datasets when compared with other multi-frame optical flow methods while achieving highly comparable results with state-of-the-art two-frame methods. We believe that, in future, with the availability of larger training datasets with multiple input sequences from various scenes, multi-frame based methods can achieve a more generalized and accurate optical flow estimates when compared with two-frame methods.

\section*{Acknowledgement}
This research was supported in part by a \textit{Herff graduate fellowship} from the Herff College of Engineering at The University of Memphis to one of the authors (Fisseha A Ferede).


\pagebreak
\clearpage
\bibliographystyle{plainnat}
\bibliography{References.bib}
\end{document}